\documentclass[11pt]{article}
\usepackage{array}
\usepackage{acl2014}
\usepackage{times}
\usepackage{url}
\usepackage{latexsym}
\usepackage{amsmath}
\usepackage{amssymb}
\usepackage{dsfont}
\usepackage{multirow}
\usepackage{balance}
\usepackage[hang,flushmargin]{footmisc} 
\usepackage{graphicx}
\usepackage[small]{caption}
\usepackage{multirow}
\usepackage[belowskip=-5pt, aboveskip=5pt]{caption}
\usepackage{pifont}
\usepackage{url}
\title{Convolutional Neural Networks for Sentence Classification}

\author{Yoon Kim \\ New York University \\ {\tt yhk255@nyu.edu}}

\date{}

\begin{document}
\maketitle
\begin{abstract}
We report on a series of experiments with convolutional neural networks (CNN) trained on top of pre-trained word vectors for sentence-level classification tasks. We show that a simple CNN with little hyperparameter tuning and static vectors achieves excellent results on multiple benchmarks. Learning task-specific vectors through fine-tuning offers further gains in performance. We additionally propose a simple modification to the architecture to allow for the use of both task-specific and static vectors. The CNN models discussed herein improve upon the state of the art on 4 out of 7 tasks, which include sentiment analysis and question classification.
\end{abstract}

\section{Introduction}
Deep learning models have achieved remarkable results in computer vision \cite{Krizhevsky:2012} and speech recognition \cite{Graves:2013} in recent years. Within natural language processing, much of the  work with deep learning methods has involved learning word vector representations through neural language models (Bengio et al., 2003; Yih et al., 2011; Mikolov et al., 2013) and performing composition over the learned word vectors for classification \cite{Collobert:2011}. Word vectors, wherein words are projected from a sparse, 1-of-$V$ encoding (here $V$ is the vocabulary size) onto a lower dimensional vector space via a hidden layer, are essentially feature extractors that encode semantic features of words in their  dimensions. In such dense representations, semantically close words are likewise close---in euclidean or cosine distance---in the lower dimensional vector space.

Convolutional neural networks (CNN) utilize layers with convolving filters that are applied to local features \cite{LeCun:1998}. Originally invented for computer vision, CNN models have subsequently been shown to be effective for NLP and have achieved excellent results in semantic parsing \cite{Yih:2014}, search query retrieval \cite{Shen:2014}, sentence modeling \cite{Kalch:2014}, and other traditional NLP tasks \cite{Collobert:2011}. 

In the present work, we train a simple CNN with one layer of convolution on top of word vectors obtained from an unsupervised neural language model. These vectors were trained by Mikolov et al. \shortcite{Mikolov:2013} on 100 billion words of Google News, and are publicly available.\footnote{\url{https://code.google.com/p/word2vec/}} We initially keep the word vectors static and learn only the other parameters of the model. Despite little tuning of hyperparameters, this simple model achieves excellent results on multiple benchmarks, suggesting that the pre-trained vectors are `universal' feature extractors that can be utilized for various classification tasks. Learning task-specific vectors through fine-tuning results in further improvements. We finally describe a simple modification to the architecture to allow for the use of both pre-trained and task-specific vectors by having multiple channels.

Our work is philosophically similar to Razavian et al. \shortcite{Razavian:2014} which showed that for image classification, feature extractors obtained from a pre-trained deep learning model perform well on a variety of tasks---including tasks that are very different from the original task for which the feature extractors were trained.

\section{Model}
The model architecture, shown in figure 1, is a slight variant of the CNN architecture of Collobert et al. \shortcite{Collobert:2011}. Let $\mathbf{x}_i \in \mathbb{R}^{k}$ be the $k$-dimensional word vector corresponding to the $i$-th word in the sentence. A sentence of length $n$ (padded where necessary) is represented as
\begin{figure*}
  \center
  \includegraphics[width=16cm,height=12cm]{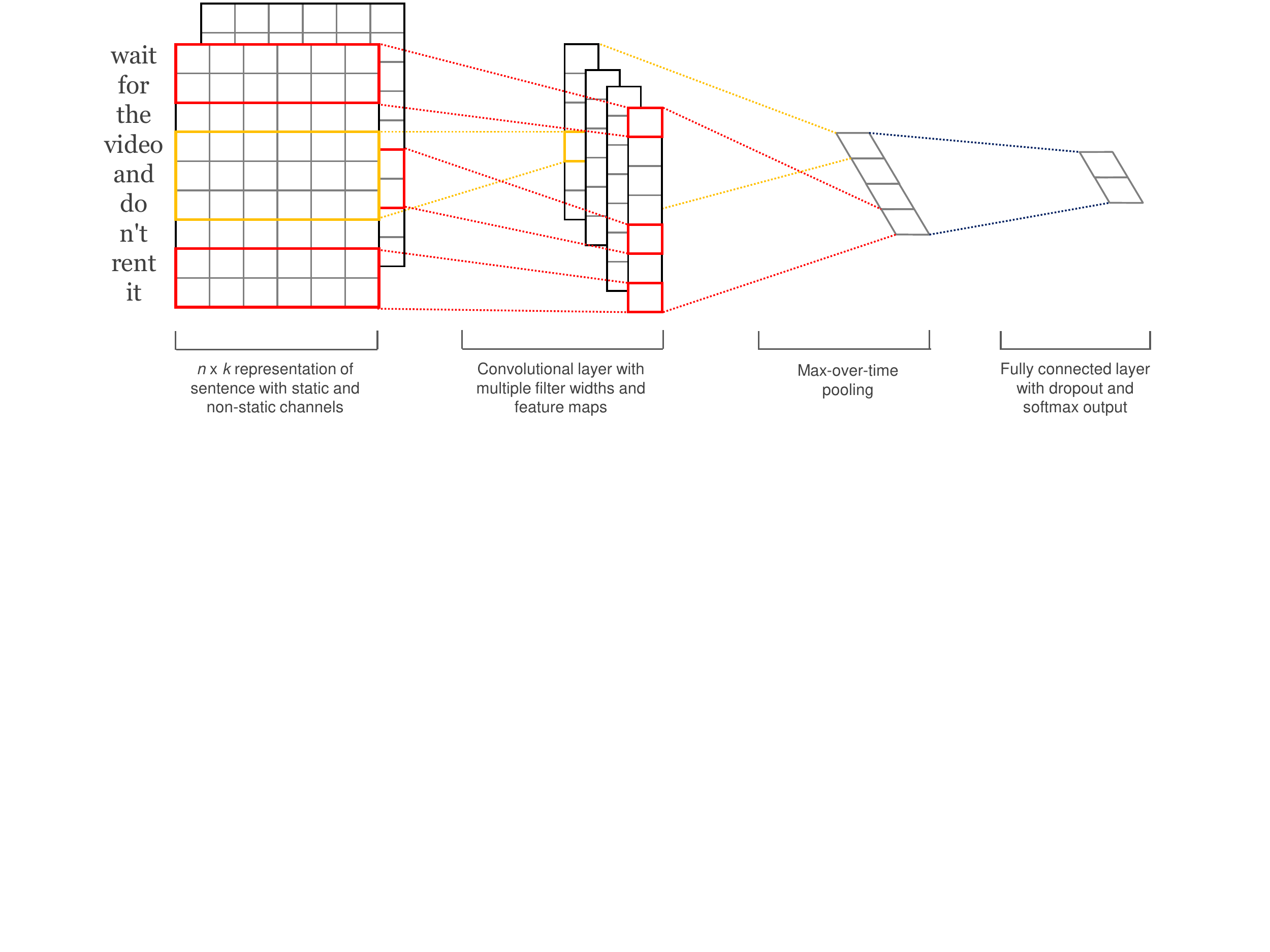}
\vspace{-6.7cm}
  \caption{Model architecture with two channels for an example sentence.}
\end{figure*}

\begin{equation}
\mathbf{x}_{1:n} = \mathbf{x}_1 \oplus \mathbf{x}_2 \oplus \ldots \oplus \mathbf{x}_n,
\end{equation}
where $\oplus$ is the concatenation operator. In general, let $\mathbf{x}_{i:i+j}$ refer to the concatenation of words $\mathbf{x}_i, \mathbf{x}_{i+1}, \ldots, \mathbf{x}_{i+j}$.
A convolution operation involves a \emph{filter} $\mathbf{w} \in \mathbb{R}^{hk}$, which is applied to a window of $h$ words to produce a new feature. For example, a feature $c_i$ is generated from a window of words $\mathbf{x}_{i:i+h-1}$ by
\begin{equation}
c_i = f(\mathbf{w} \cdot \mathbf{x}_{i:i+h-1} + b).
\end{equation}
Here $b \in \mathbb{R}$ is a bias term and $f$ is a non-linear function such as the hyperbolic tangent. This filter is applied to each possible window of words in the sentence $\{\mathbf{x}_{1:h}, \mathbf{x}_{2:h+1}, \ldots, \mathbf{x}_{n-h+1:n}\}$ to produce a \emph{feature map}
\begin{equation}
\mathbf{c} = [c_1, c_2, \ldots, c_{n-h+1}],
\end{equation}
with $\mathbf{c} \in \mathbb{R}^{n-h+1}$. We then apply a max-over-time pooling operation \cite{Collobert:2011} over the feature map and take the maximum value $\hat{c} = \max \{\mathbf{c}\}$ as the feature corresponding to this particular filter. The idea is to capture the most important feature---one with the highest value---for each feature map. This pooling scheme naturally deals with variable sentence lengths. 

We have described the process by which $one$ feature is extracted from $one$ filter. The model uses multiple filters (with varying window sizes) to obtain multiple features. These features form the penultimate layer and are passed to a fully connected softmax layer whose output is the probability distribution over labels.

In one of the model variants, we experiment with having two `channels' of word vectors---one that is kept static throughout training and one that is fine-tuned via backpropagation  (section 3.2).\footnote{We employ language from computer vision where a color image has red, green, and blue channels.} In the multichannel architecture, illustrated in figure 1, each filter is applied to both channels and the results are added to calculate $c_i$ in equation (2). The model is otherwise equivalent to the single channel architecture.
\subsection{Regularization}
For regularization we employ dropout on the penultimate layer with a constraint on $l_2$-norms of the weight vectors \cite{Hinton:2012}. Dropout prevents co-adaptation of hidden units by randomly dropping out---i.e., setting to zero---a proportion $p$ of the hidden units during foward-backpropagation. That is, given the penultimate layer $\mathbf{z} = [\hat{c}_1, \ldots, \hat{c}_m]$ (note that here we have $m$ filters), instead of using
\begin{equation}
y = \mathbf{w} \cdot \mathbf{z} + b
\end{equation}
for output unit $y$ in forward propagation, dropout uses
\begin{equation}
y = \mathbf{w} \cdot (\mathbf{z}\circ\mathbf{r}) + b,
\end{equation}
where $\circ$ is the element-wise multiplication operator and $\mathbf{r} \in \mathbb{R}^m$ is a `masking' vector of Bernoulli random variables with probability $p$ of being 1. Gradients are backpropagated only through the unmasked units. At test time, the learned weight vectors are scaled by $p$ such that $\hat{\mathbf{w}} = p\mathbf{w}$, and $\hat{\mathbf{w}}$ is used (without dropout) to score unseen sentences. We additionally constrain $l_2$-norms of the weight vectors by rescaling $\mathbf{w}$ to have $||\mathbf{w}||_2 = s$ whenever $||\mathbf{w}||_2 > s$ after a gradient descent step.

\section{Datasets and Experimental Setup}
We test our model on various benchmarks. Summary statistics of the datasets are in table 1.
\begin{table}
\center
\begin{tabular}{ >{\centering\arraybackslash}m{1cm} || >{\centering\arraybackslash}m{0.15cm} | >{\centering\arraybackslash}m{0.25cm} | >{\centering\arraybackslash}m{0.825cm} | >{\centering\arraybackslash}m{0.825cm} | >{\centering\arraybackslash}m{0.825cm} | >{\centering\arraybackslash}m{0.8cm} }
\hline
\textbf{Data} & $c$ & $l$ & $N$ & $|V|$ & $|V_{pre}|$ & \emph{Test} \\ \hline
MR & $2$ & $20$ & $10662$ &  $18765$ & $16448$ & CV \\ 
SST-1 &  $5$ & $18$ &  $11855$ &  $17836$ & $16262$ & $2210$ \\ 
SST-2 &  $2$ & $19$ &  $9613$ &  $16185$ & $14838$ & $1821$ \\ 
Subj & $2$ & $23$ & $10000$ & $21323$& $17913$ & CV \\
TREC & $6$ & $10$ & $5952$ & $9592$ & $9125$ & $500$ \\ 
CR &  $2$ & $19$ & $3775$ &  $5340$ & $5046$ & CV \\
MPQA & $2$ & $3$ &  $10606$ & $6246$ & $6083$  & CV \\ \hline
\end{tabular}
\caption{Summary statistics for the datasets after tokenization. $c$: Number of target classes. $l$: Average sentence length. $N$: Dataset size. $|V|$: Vocabulary size. $|V_{pre}|$: Number of words present in the set of pre-trained word vectors. \emph{Test}: Test set size (CV means there was no standard train/test split and thus 10-fold CV was used).}
\end{table}
\begin{itemize}
  \item \textbf{MR}: Movie reviews with one sentence per review. Classification involves detecting positive/negative reviews \cite{Pang:2005}.\footnote{https://www.cs.cornell.edu/people/pabo/movie-review-data/}
 \item \textbf{SST-1}: Stanford Sentiment Treebank---an extension of MR but with train/dev/test splits provided and fine-grained labels (very  positive, positive, neutral, negative, very negative), re-labeled by Socher et al. \shortcite{Socher:2013}.\footnote{\url{http://nlp.stanford.edu/sentiment/} Data is actually provided at the phrase-level and hence we train the model on both phrases and sentences but only score on sentences at test time, as in Socher et al. \shortcite{Socher:2013}, Kalchbrenner et al. \shortcite{Kalch:2014}, and Le and Mikolov \shortcite{Le:2014}. Thus the training set is an order of magnitude larger than listed in table 1.}
 \item \textbf{SST-2}: Same as SST-1 but with neutral reviews removed and binary labels.
 \item \textbf{Subj}: Subjectivity dataset where the task is to classify a sentence as being subjective or objective \cite{Pang:2004}.
  \item \textbf{TREC}: TREC question dataset---task involves classifying a question into 6 question types (whether the question is about person, location, numeric information, etc.) \cite{Li:2002}.\footnote{\url{http://cogcomp.cs.illinois.edu/Data/QA/QC/}}
  \item \textbf{CR}: Customer reviews of various products (cameras, MP3s etc.). Task is to predict positive/negative reviews \cite{Hu:2004}.\footnote{\url{http://www.cs.uic.edu/~liub/FBS/sentiment-analysis.html}}
  \item \textbf{MPQA}: Opinion polarity detection subtask of the MPQA dataset \cite{Wiebe:2005}.\footnote{\url{http://www.cs.pitt.edu/mpqa/}}
\end{itemize}
\subsection{Hyperparameters and Training}
For all datasets we use: rectified linear units, filter windows ($h$) of 3, 4, 5 with 100 feature maps each, dropout rate ($p$) of 0.5, $l_2$ constraint ($s$) of 3, and mini-batch size of 50. These values were chosen via a grid search on the SST-2 dev set.

We do not otherwise perform any dataset-specific tuning other than early stopping on dev sets. For datasets without a standard dev set we randomly select 10\% of the training data as the dev set. Training is done through stochastic gradient descent over shuffled mini-batches with the Adadelta update rule \cite{Zeiler:2012}.
\subsection{Pre-trained Word Vectors}
Initializing word vectors with those obtained from an unsupervised neural language model is a popular method to improve performance in the absence of a large supervised training set (Collobert et al., 2011; Socher et al., 2011; Iyyer et al., 2014). We use the publicly available \texttt{word2vec} vectors that were trained on 100 billion words from Google News. The vectors have dimensionality of 300 and were trained using the continuous bag-of-words architecture \cite{Mikolov:2013}. Words not present in the set of pre-trained words are initialized randomly.
\begin{table*}[ht]
\center
\begin{tabular}{>{\arraybackslash}m{5.95cm} ||>{\centering\arraybackslash}m{0.9cm}|>{\centering\arraybackslash}m{0.95cm}| >{\centering\arraybackslash}m{0.95cm}| >{\centering\arraybackslash}m{0.9cm}| >{\centering\arraybackslash}m{0.9cm}| >{\centering\arraybackslash}m{0.9cm}| >{\centering\arraybackslash}m{0.9cm}  }
\hline
\textbf{Model} & MR & SST-1 & SST-2 & Subj & TREC & CR & MPQA \\ \hline
CNN-rand & $76.1$ & $45.0$ & $82.7$ & $89.6$ & $91.2$ & $79.8$& $83.4$\\ 
CNN-static & $81.0$ & $45.5$ & $86.8$ & $93.0$ & $92.8$ & $84.7$& $\boldsymbol{89.6}$\\ 
CNN-non-static & $\boldsymbol{81.5}$ & $48.0$ & $87.2$ & $93.4$ & $93.6$ & $84.3$& $89.5$\\ 
CNN-multichannel & $81.1$ & $47.4$ & $\boldsymbol{88.1}$ & $93.2$ & $92.2$ & $\boldsymbol{85.0}$& $89.4$\\  \hline
RAE \cite{Socher:2011} & $77.7$ & $43.2$ & $82.4$ & $-$ & $-$ & $-$ & $86.4$\\
MV-RNN \cite{Socher:2012} & $79.0$ & $44.4$ & $82.9$ & $-$ & $-$& $-$& $-$ \\
RNTN \cite{Socher:2013} & $-$ & $45.7$ & $85.4$ & $-$ & $-$& $-$& $-$\\
DCNN \cite{Kalch:2014} & $-$ & $48.5$& $86.8$ & $-$ & $93.0$ & $-$& $-$ \\
Paragraph-Vec \cite{Le:2014} & $-$ & $\boldsymbol{48.7}$& $87.8$ & $-$ & $-$ & $-$& $-$ \\
CCAE \cite{Hermann:2013} & $77.8$ & $-$ & $-$ & $-$ & $-$ & $-$& $87.2$\\
Sent-Parser \cite{Dong:2014} & $79.5$ & $-$ & $-$ & $-$ & $-$ & $-$ & $86.3$\\
NBSVM \cite{Wang:2012} & $79.4$ & $-$ & $-$ & $93.2$ & $-$ & $81.8$& $86.3$ \\
MNB \cite{Wang:2012} & $79.0$ & $-$ & $-$ & $\boldsymbol{93.6}$ & $-$ & $80.0$& $86.3$ \\
G-Dropout \cite{Wang:2013} & $79.0$ & $-$ & $-$ & $93.4$ & $-$ & $82.1$& $86.1$ \\
F-Dropout \cite{Wang:2013} & $79.1$ & $-$ & $-$ & $\boldsymbol{93.6}$ & $-$ & $81.9$& $86.3$ \\
Tree-CRF \cite{Nakagawa:2010} & $77.3$ & $-$ & $-$ & $-$ & $-$ & $81.4$& $86.1$ \\
CRF-PR \cite{Yang:2014} & $-$ & $-$ & $-$ & $-$ & $-$ & $82.7$& $-$ \\
SVM$_{S}$ \cite{Silva:2011} & $- $ & $-$ & $-$ & $-$ & $\boldsymbol{95.0}$ & $-$ & $-$ \\ \hline
\end{tabular}
\caption{Results of our CNN models against other methods. \textbf{RAE}: Recursive Autoencoders with pre-trained word vectors from Wikipedia \cite{Socher:2011}. \textbf{MV-RNN}: Matrix-Vector Recursive Neural Network with parse trees \cite{Socher:2012}. \textbf{RNTN}: Recursive Neural Tensor Network with tensor-based feature function and parse trees \cite{Socher:2013}. \textbf{DCNN}: Dynamic Convolutional Neural Network with k-max pooling \cite{Kalch:2014}. \textbf{Paragraph-Vec}: Logistic regression on top of paragraph vectors \cite{Le:2014}. \textbf{CCAE}: Combinatorial Category Autoencoders with combinatorial category grammar operators \cite{Hermann:2013}.  \textbf{Sent-Parser}: Sentiment analysis-specific parser \cite{Dong:2014}. \textbf{NBSVM, MNB}: Naive Bayes SVM and Multinomial Naive Bayes with uni-bigrams from Wang and Manning \shortcite{Wang:2012}.  \textbf{G-Dropout, F-Dropout}: Gaussian Dropout and Fast Dropout from Wang and Manning \shortcite{Wang:2013}. \textbf{Tree-CRF}: Dependency tree with Conditional Random Fields \cite{Nakagawa:2010}. \textbf{CRF-PR}: Conditional Random Fields with Posterior Regularization \cite{Yang:2014}. \textbf{SVM$_{S}$}: SVM with uni-bi-trigrams, wh word, head word, POS, parser, hypernyms, and 60 hand-coded rules as features from Silva et al. \shortcite{Silva:2011}.}
\end{table*}
\subsection{Model Variations}
We experiment with several variants of the model.
\begin{itemize}
\item \textbf{CNN-rand}: Our baseline model where all words are randomly initialized and then modified during training.
\item \textbf{CNN-static}: A model with pre-trained vectors from \texttt{word2vec}. All words---including the unknown ones that are randomly initialized---are kept static and only the other parameters of the model are learned.
\item \textbf{CNN-non-static}: Same as above but the pre-trained vectors are fine-tuned for each task.
\item \textbf{CNN-multichannel}: A model with two sets of word vectors. Each set of vectors is treated as a `channel' and each filter is applied to both channels, but gradients are backpropagated only through one of the channels. Hence the model is able to fine-tune one set of vectors while keeping the other static. Both channels are initialized with \texttt{word2vec}.
\end{itemize}
In order to disentangle the effect of the above variations versus other random factors, we eliminate other sources of randomness---CV-fold assignment, initialization of unknown word vectors, initialization of CNN parameters---by keeping them uniform within each dataset.

\section{Results and Discussion}
Results of our models against other methods are listed in table 2. Our baseline model with all randomly initialized words (CNN-rand) does not perform well on its own. While we had expected performance gains through the use of pre-trained vectors, we were surprised at the magnitude of the gains. Even a simple model with static vectors (CNN-static) performs remarkably well, giving competitive results against the more sophisticated deep learning models that utilize complex pooling schemes \cite{Kalch:2014} or require parse trees to be computed beforehand \cite{Socher:2013}.  These results suggest that the pre-trained vectors are good, `universal' feature extractors and can be utilized across datasets. Fine-tuning the pre-trained vectors for each task gives still further improvements (CNN-non-static).
\subsection{Multichannel vs. Single Channel Models}
We had initially hoped that the multichannel architecture would prevent overfitting (by ensuring that the learned vectors do not deviate too far from the original values) and thus work better than the single channel model, especially on smaller datasets. The results, however, are mixed, and further work on regularizing the fine-tuning process is warranted. For instance, instead of using an additional channel for the non-static portion, one could maintain a single channel but employ extra dimensions that are allowed to be modified during training.

\subsection{Static vs. Non-static Representations}
As is the case with the single channel non-static model, the multichannel model is able to fine-tune the non-static channel to make it more specific to the task-at-hand. For example, \textit{good} is most similar to \textit{bad} in \texttt{word2vec}, presumably because they are (almost) syntactically equivalent. But for vectors in the non-static channel that were fine-tuned on the SST-2 dataset, this is no longer the case (table 3). Similarly,  \textit{good} is arguably closer to \textit{nice} than it is to \textit{great} for expressing sentiment, and this is indeed reflected in the learned vectors.

For (randomly initialized) tokens not in the set of pre-trained vectors, fine-tuning allows them to learn more meaningful representations: the network learns that exclamation marks are  associated with effusive expressions and that commas are conjunctive (table 3).
\begin{table}
\centering
\begin{tabular}{c|c|c} \hline
\multirow{2}{*}{} & \multicolumn{2}{c}{Most Similar Words for} \\ \cline{2-3} & Static Channel & Non-static Channel \\ \hline
\multirow{4}{*}{\it{\textbf{bad}}} & \it{good} & \it{terrible} \\ & \it{terrible} & \it{horrible} \\ & \it{horrible} & \it{lousy} \\ & \it{lousy} & \it{stupid} \\ \hline
\multirow{4}{*}{\it{\textbf{good}}} & \it{great} & \it{nice} \\ & \it{bad} & \it{decent} \\ & \it{terrific} & \it{solid} \\ & \it{decent} & \it{terrific} \\ \hline
\multirow{4}{*}{\it{\textbf{n't}}} & \it{os} & \it{not} \\ & \it{ca} & \it{never} \\ & \it{ireland} & \it{nothing} \\ & \it{wo} & \it{neither} \\ \hline
\multirow{4}{*}{\it{\textbf{!}}} & \it{2,500} & \it{2,500} \\ & \it{entire} & \it{lush} \\ & \it{jez} & \it{beautiful} \\ & \it{changer} & \it{terrific} \\ \hline
\multirow{4}{*}{\it{\textbf{,}}} & \it{decasia} & \it{but} \\ & \it{abysmally} & \it{dragon} \\ & \it{demise} & \it{a} \\  & \it{valiant} & \it{and} \\ \hline
\end{tabular}
\caption{Top 4 neighboring words---based on cosine similarity---for vectors in the static channel (left) and fine-tuned vectors in the non-static channel (right) from the multichannel model on the SST-2 dataset after training.}
\end{table}
\subsection{Further Observations}
We report on some further experiments and observations:
\begin{itemize}
\item Kalchbrenner et al. \shortcite{Kalch:2014} report much worse results with a CNN that has essentially the same architecture as our single channel model. For example, their Max-TDNN (Time Delay Neural Network) with randomly initialized words obtains $37.4\%$ on the SST-1 dataset, compared to $45.0\%$ for our model. We attribute such discrepancy to our CNN having much more capacity (multiple filter widths and feature maps).
\item Dropout proved to be such a good regularizer that it was fine to use a larger than necessary network and simply let dropout regularize it. Dropout consistently added 2\%--4\% relative performance.
\item When randomly initializing words not in \texttt{word2vec}, we obtained slight improvements by sampling each dimension from $U[-a,a]$ where $a$ was chosen such that the randomly initialized vectors have the same variance as the pre-trained ones. It would be interesting to see if employing more sophisticated methods to mirror the distribution of pre-trained vectors in the initialization process gives further improvements.
\item We briefly experimented with another set of publicly available word vectors trained by Collobert et al. \shortcite{Collobert:2011} on Wikipedia,\footnote{\url{http://ronan.collobert.com/senna/}} and found that \texttt{word2vec} gave far superior performance. It is not clear whether this is due to Mikolov et al. \shortcite{Mikolov:2013}'s architecture or the 100 billion word Google News dataset.
\item Adadelta \cite{Zeiler:2012} gave similar results to Adagrad \cite{Duchi:2011} but required fewer epochs.
\end{itemize}

\section{Conclusion}
In the present work we have described a series of experiments with convolutional neural networks built on top of \texttt{word2vec}. Despite little tuning of hyperparameters, a simple CNN with one layer of convolution performs remarkably well. Our results add to the well-established evidence that unsupervised pre-training of word vectors is an important ingredient in deep learning for NLP.
\subsection*{Acknowledgments}
We would like to thank Yann LeCun and the anonymous reviewers for their helpful feedback and suggestions.
\balance

\end{document}